\documentclass[%
 aip,
 amsmath,amssymb,
 reprint,%
]{revtex4-1}

\usepackage{graphicx}% Include figure files
\usepackage{dcolumn}% Align table columns on decimal point
\usepackage{bm}% bold math
\usepackage[utf8]{inputenc}
\usepackage[T1]{fontenc}
\usepackage{mathptmx}
\usepackage{etoolbox}

\makeatletter
\def\@email#1#2{%
 \endgroup
 \patchcmd{\titleblock@produce}
  {\frontmatter@RRAPformat}
  {\frontmatter@RRAPformat{\produce@RRAP{*#1\href{mailto:#2}{#2}}}\frontmatter@RRAPformat}
  {}{}
}%
\makeatother
\begin{document}

\preprint{AIP/123-QED}

\title[Internal noise in deep and recurrent neural networks helps with learning]{Internal noise in hardware deep and recurrent neural networks helps with learning}
% Force line breaks with \\
\author{I.D. Kolesnikov}%
 \email{kole200@yandex.ru}
\author{N. Semenova}%
 \email{semenovani@sgu.ru}
 \affiliation{Saratov State University, Astrakhanskaya str. 83, Saratov 410012, Russia}%

\date{\today}% It is always \today, today,

\begin{abstract}

Recently, the field of hardware neural networks has been actively developing, where neurons and their connections are not simulated on a computer but are implemented at the physical level, transforming the neural network into a tangible device. In this paper, we investigate how internal noise during the training of neural networks affects the final performance of recurrent and deep neural networks. We consider feedforward networks (FNN) and echo state networks (ESN) as examples. The types of noise examined originated from a real optical implementation of a neural network. However, these types were subsequently generalized to enhance the applicability of our findings on a broader scale. The noise types considered include additive and multiplicative noise, which depend on how noise influences each individual neuron, and correlated and uncorrelated noise, which pertains to the impact of noise on groups of neurons (such as the hidden layer of FNNs or the reservoir of ESNs). In this paper, we demonstrate that, in most cases, both deep and echo state networks benefit from internal noise during training, as it enhances their resilience to noise. Consequently, the testing performance at the same noise intensities is significantly higher for networks trained with noise than for those trained without it. Notably, only multiplicative correlated noise during training has minimal has almost no impact on both deep and recurrent networks.

\end{abstract}

\maketitle

\begin{quotation}
Over the past few years, artificial neural networks (ANNs) have found their application in solving many problems. In terms of computation, ANN modeling is a very resource-intensive task. Despite the existence of high-power computing clusters with the ability to parallelize computations, modeling a neural network on digital equipment is a bottleneck in network scaling, speed of receiving or processing information and energy efficiency. In recent years, more and more researchers in the field of neural networks are interested in creating hardware networks in which neurons and the connection between them represent a real device capable of learning and solving problems. However, experimental and hardware setups always contain noise of various natures, so studying their influence is a pressing problem, the solution of which will help improve the efficiency of training in the presence of noise. The study of the influence of various types of noise within the framework of machine learning is aimed at determining the properties of noise effects that can be accumulated by a neural network, or, conversely, be suppressed by a network itself. 

% In this regard, a number of questions arise regarding the influence of specific features and characteristics of noise exposure on the learning process and the functioning of neural networks. The practical significance of such problems is explained by the need to create artificial neural networks that are resistant to stochastic disturbances. Issues of noise influence are especially important in the context of developing real experimental prototypes of trained artificial neural networks, which are always exposed to both internal noise sources and external random influences.
\end{quotation}

\section{Introduction}\label{sec:intro}
Over the past few years, artificial neural networks (ANNs) have found their application in solving many problems \cite{Lecun2015}. These problems include image recognition \cite{Krizhevsky2017, Maturana2015}, their classification, improvement of audio recordings, speech recognition \cite{Graves2013}, prediction of climate events \cite{Kar2009} and many others. 

As ANNs and their tasks become more and more complex, we may soon face a crisis situation \cite{Hasler2013,Gupta2015}, namely, that the tasks will already be so complex that the capabilities of modern computers and computing clusters will not be sufficient to meet the growing needs. Despite the existence of high-power computing clusters with the ability to parallelize computations, modeling a neural network on digital equipment is a bottleneck in network scaling, speed of receiving, processing information and energy efficiency. Here a new direction in the design of ANNs comes to the rescue -- hardware neural networks \cite{Karniadakis2021} (analog neural network in some literature). This direction involves the creation of neural networks in hardware, when the neurons and the connection between them are implemented at the physical level, which allows for a significant increase in speed and energy efficiency \cite{Aguirre2024,Chen2023}. According to this approach, neural networks are not created using a computer, but are a real device capable of learning and solving problems. The neurons themselves and the connections between them are implemented at the physical level, i.e. the network is not modeled on a computer, but is implemented in hardware according to the corresponding physical principles. In recent years, there has been an exponential growth in work with hardware implementations of ANNs. The most effective ANNs at the moment are those based on lasers \cite{Brunner2013a}, memristors \cite{Tuma2016}, and spin-torque oscillators \cite{Torrejon2017}. The connection between neurons in optical implementations of ANNs is based on the principles of holography \cite{Psaltis1990}, diffraction \cite{Bueno2018, Lin2018}, integrated networks of Mach-Zehnder modulators \cite{Shen2017}, spectral channel multiplexing \cite{Tait2017}, and optical connections implemented using a 3D printer \cite{Moughames2020,Dinc2020,Moughames2020a}.

Hardware networks provide a significant advantage in terms of speed, but they may have internal noise coming from the components of the experimental setup. In this case, the internal noise of individual components can propagate throughout the network and potentially accumulate. Therefore, such networks may be vulnerable to noise, which is a significant obstacle to the training and operation process. At the moment, several works have been published showing that the presence of noise inside an experimentally implemented ANN interferes with training or operation \cite{Frye1991,Dibazar2006,Soriano2015,Moon2019,Janke2020}. In computer-simulated networks, noise was usually added only to the input data, which affected the operation and training process \cite{Shen2017,Tait2017}. At the moment, it is known that noise in the input data can have a positive effect on the training process, helping the network to get out of a local minimum during gradient descent, and there are special network training algorithms based on the input noise \cite{Audhkhasi2016, Zhou2019}.

In terms of hardware neural networks, it is more important to consider not only the effect of noise on the input signal, but also the effect of internal noise coming from various network components (for example, neurons and their connections). This will allow us to understand which types of noise can really be critical for certain types of networks, and which can be suppressed by the network itself during the training process. In our previous works, we have already considered the effect of noise on a trained deep \cite{Semenova2022NN} and recurrent \cite{Semenova2025echo} neural networks, and also proposed strategies for reducing various types of noise \cite{Semenova2022Chaos, Semenova2024Chaos}. The key feature of the present article is that here we consider how internal noise affects the training process of deep and recurrent networks and compare test accuracy under noise exposure for a network trained with noise and networks trained with different noise intensities.

%The general structure of the paper is as follows. After this section, there is a section \ref{sec:system} describing the networks that will be considered. In this paper, these are the feedforward network (Sec.~\ref{sec:system:FNN}) and the echo network (Sec.~\ref{sec:system:ESN}). Then, we introduce the classification of internal noise and describe how noisy influence is included (Sec.~\ref{sec:system:noise}). Next, we describe in detail how additive and multiplicative, correlated and uncorrelated noise affects the training and performance of feedforward neural network (Sec.~\ref{sec:noise_FNN}) and echo state network (Sec.~\ref{sec:noise_ESN}).

\section{System under study}\label{sec:system}
In this article, we consider a deep network using the example of feedforward neural network (FNN) and a recurrent network using the example of echo state network (ESN).
\subsection{Feedforward neural network}\label{sec:system:FNN}
\begin{figure}[h]
\includegraphics[width=\linewidth]{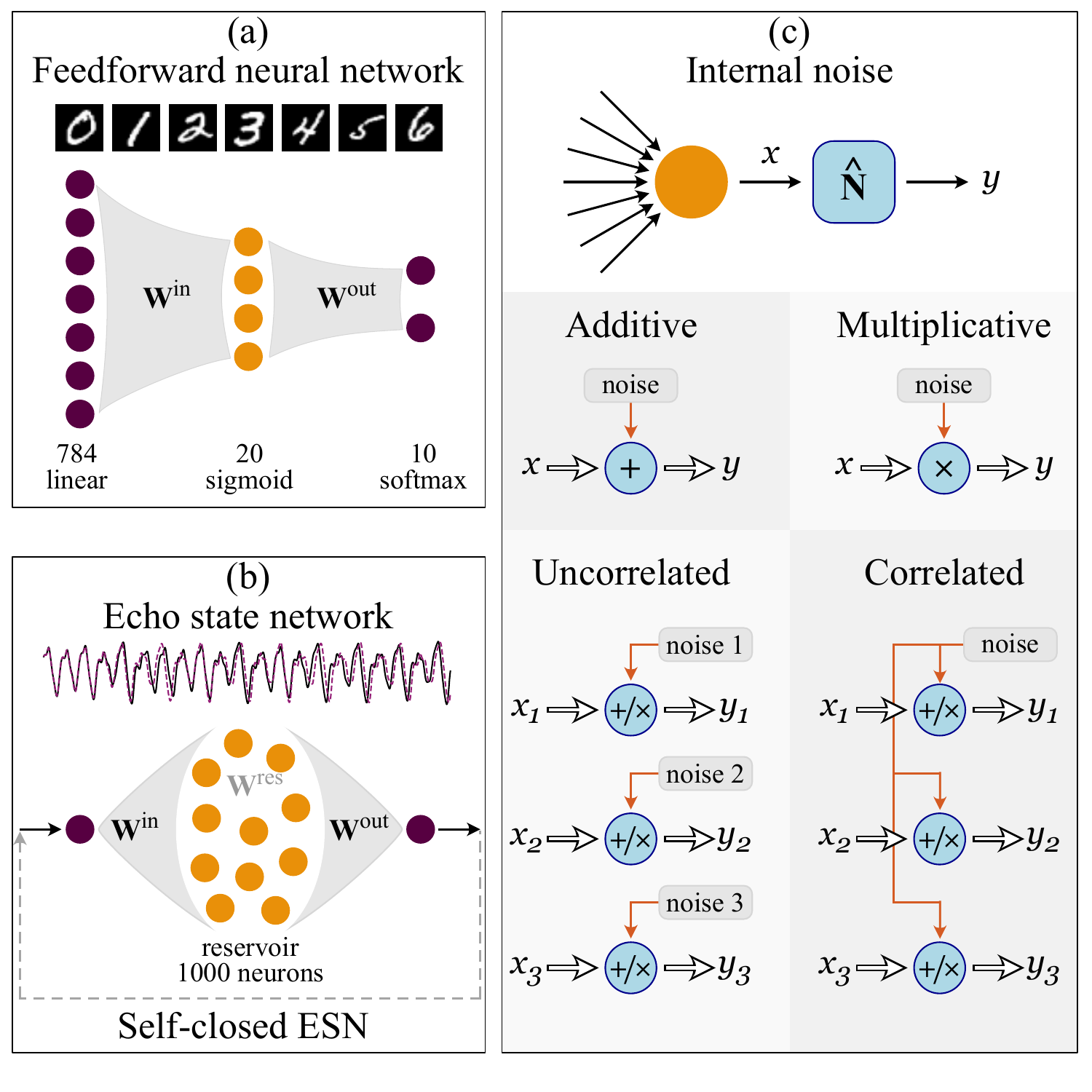}
\caption{\label{fig:scheme} Schematic representation of considered FNN (a), ESN (b) and how internal noise is introduced into artificial neurons (c).}
\end{figure}

The considered FNN is shown schematically in Fig.~\ref{fig:scheme}(a). This network consists of three layers and is trained to recognize handwritten digits from MNIST database \cite{LeCun1998}. This database has 60,000 training and 10,000 testing images of size 28$\times$28 pixels in a gray scale. The network's input data is therefore transformed to be a vector of length 784. The value of each pixel is normalized to be a real number in the range 0 to 1 (instead of the original integers in the range 0 to 255), and then it is sent to the input of the corresponding input neuron. Based on the solution of the classification problem for 10 classes (numbers 0--9), the output layer should have 10 neurons with the softmax activation function. In this case, the network's response to the incoming image is not the output signal of the output neurons itself, but which output neuron has the largest output.

This network has one hidden layer consisting of 20 artificial neurons with sigmoid activation function $f(x) = 1/(1+e^{-x})$. These three layers are connected with each other via connection matrices $\mathbf{W}^\mathrm{in}$ and $\mathbf{W}^\mathrm{out}$ of size $(784\times 20)$ and $(20\times 10)$, respectively. Then the output signal of hidden neurons can be set as
\begin{equation}
x^\mathrm{hidd}_i = f\Big( \sum\limits_{j=0}^{784}x^\mathrm{in}_j W^\mathrm{in}_{j,i} \Big) \ \ \ \ \text{or} \ \ \ \mathbf{x}^\mathrm{hidd} = f( \mathbf{x}^\mathrm{in} \mathbf{W}^\mathrm{in} ),
\end{equation}
where $\mathbf{x}^\mathrm{in}$ is the vector of signals from the neurons of the input layer, $\mathbf{x}^\mathrm{hidd}$ is the vector of signals from the neurons of the hidden layer. Further we will add the internal noise into hidden layer of FNN, and we will denote the signal of noisy neurons as $\mathbf{y}^\mathrm{hidd}$. Therefore, the output signal of FNN is
\begin{equation}
\begin{array}{c}
x^\mathrm{out}_i = \mathrm{softmax}( \sum\limits_{j=0}^{20}y^\mathrm{hidd}_j W^\mathrm{out}_{j,i} ) \ \ \ \ \text{or} \\
\mathbf{x}^\mathrm{out} = \mathrm{softmax}( \mathbf{y}^\mathrm{hidd} \mathbf{W}^\mathrm{out} ),
\end{array}
\end{equation}
where $\mathbf{x}^\mathrm{out}$ is the vector of signals from the neurons of the output layer. In the noise free FNN, $\mathbf{y}^\mathrm{hidd}=\mathbf{x}^\mathrm{hidd}$.

The network is trained using gradient descent and back propagation. In order to compare the performance of FNNs for different noise types, further we will compare the accuracies of digits recognition during training the FNN and after training on testing data.

\subsection{Echo state neural network}\label{sec:system:ESN}

The second type of considered network is recurrent neural network, This network is trained to predict the behavior of a chaotic Mackey--Glass system. The last is defined by a single first-order differential equation with a time delayed feedback:
\begin{equation}\label{eq:MG}
\frac{du}{dt} = \beta\frac{u_\tau}{1+u^n_\tau} - \gamma u,
\end{equation}
where $u$ is the variable of system, which realization $u(t)$ is predicted, $u_\tau=u(t-\tau)$ is its delayed state, parameters $\beta=0.2$, $\gamma=0.1$, $\tau=17$, $n=10$ are fixed throughout the paper and correspond to common chaotic dynamics of Mackey--Glass system.

The first 20,000 points of obtained realization are used for training, while the next 100 points are used for testing. The considered ESN trained for this purpose is shown schematically in Fig.~\ref{fig:scheme}(b). To predict the behaviour of one variable $u(t)$, the ESN should have one input neuron and one output neuron. The rest neurons forming reservoir have the delayed connectivity inside reservoir. The reservoir consists of $N=1000$ neurons with hyperbolic tangent activation function $g(x)=\mathrm{tanh}x$ connected with input neuron via matrix $\mathbf{W}^\mathrm{in}$ of size $(1\times N)$ and with the same reservoir at previous time step via matrix $\mathbf{W}^\mathrm{res}$ of size $(N\times N)$. Then the state of each neuron inside reservoir at time $t$ can be set as
\begin{equation}
\mathbf{x}^\mathrm{res}_t = g\Big(x^\mathrm{in}_t\mathbf{W}^\mathrm{in} + \mathbf{x}^\mathrm{res}_{t-1}\mathbf{W}^\mathrm{res} \Big).
\end{equation}
Further, the noise will be introduced only into reservoir neurons. Their output signals after the noise impact will be designated as $\mathbf{y}^\mathrm{res}$. In the case of noise free system $\mathbf{x}^\mathrm{res}=\mathbf{y}^\mathrm{res}$. The output signal of ESN can be obtained as the output signal of output neuron connected with reservoir via matrix $\mathbf{W}^\mathrm{out}$ of size $(N\times 1)$:
\begin{equation}
x^\mathrm{out}_t = \mathbf{y}^\mathrm{res}_{t}\mathbf{W}^\mathrm{out} .
\end{equation}
The signal from input layer $x^\mathrm{in}_t$ is the same as $u(t)$ from system (\ref{eq:MG}) during training and testing. If the ESN is self-closed, then the output of network $y^\mathrm{out}_t$ is used as the input of network at next time step meaning that $x^\mathrm{in}_{t+1}=x^\mathrm{out}_t$. During training the network, only $\mathbf{W}^\mathrm{out}$ is varied. The rest matrices, $\mathbf{W}^\mathrm{in}$ is set to be matrix of ones, $\mathbf{W}^\mathrm{res}$ is set randomly with spectral radius 1.2. 

In order to compare the performance of ESNs for different noise types, further we will use root mean square error (RMSE), comparing the output of ESN $y^\mathrm{out}_t$ with expected true answer $u_{t+1}$:
\begin{equation}\label{eq:rmse}
\mathrm{RMSE} = \sqrt{\frac{\sum_{t=1}^{T}(y^\mathrm{out}_t - u_{t+1})^2}{T}},
\end{equation}
where $T=100$ is the length of testing sequence.

\subsection{Internal noise}\label{sec:system:noise}
The method off introducing the internal noise is identical to our previous works \cite{Semenova2019, Semenova2022NN, Semenova2022Chaos, Semenova2025echo}. The original types of noise, their intensities and methods of introduction were obtained from a hardware implementation of an ANN in an optical experiment, proposed in Ref.~\cite{Bueno2018}. But here we consider different noise intensities in order to make results more general and applicable to other hardware networks. 

Figure \ref{fig:scheme}(c) illustrates schematically the effect of noise on one neuron and at what stage noise is introduced. A neuron receives the summed input signal from neurons of the previous layer, then the activation function is applied  to create its noise-free output state $x$. Usually this is the output signal of the neuron, but in order to include the effect of noisy analog neurons,  noise operator $\mathbf{\hat{N}}$ is applied to this signal, leading to a final noisy signal of the form $y=\mathbf{\hat{N}} x$. 

In this paper, we study the impact of noise on the hidden layer of FNN and the reservoir of ESN. Therefore, their noisy states become
\begin{equation}\label{eq:noise_operator}
y^\mathrm{hidd}_i = \mathbf{\hat{N}} x^\mathrm{hidd}_i, \ \ \  y^\mathrm{res}_i = \mathbf{\hat{N}} x^\mathrm{res}_i.
\end{equation}
Here, we describe the effect of noise using noise operator $\mathbf{\hat{N}}$ at the stage where we introduce noise. Next we will look at what this operator is depending on different types of noise. 

In a photonic experimental implementation of a neural network, it was discovered that there are two types of noise affecting a single neuron: additive noise and multiplicative noise, mathematically described by
\begin{equation}\label{eq:noise}
y_i = \mathbf{\hat{N}} x_i =  x_i\cdot\big(1+ \sqrt{2D_M}\xi_M(t,i) \big) + \sqrt{2D_A}\xi_A(t,i)
\end{equation}
Thus, additive noise (with indices `A') is added to the noise-free output signal, while the multiplicative noise (with indices `M') is multiplied on it. The notation $\xi$ corresponds to white Gaussian noise with zero mean and variance equal to unity. Its multiplier $\sqrt{2D}$ determines the overall variance equal to $2D$, and $D$ is usually refereed to as the intensity of the noise source. The subscripts `A' and `M' for variables $\xi$ and $D$ correspond to additive and multiplicative noise, respectively. Since $\xi$ has a zero mean, simply multiplying it by the desired signal can lead to losing the entire signal, hence multiplicative noise is introduced by multiplication with $(1+\xi)$.

The previous provides a classification of noise depending on the effect on one isolated neuron. Now let us consider at the types that describe the effect on a group of neurons. As in our previous works \cite{Semenova2019, Semenova2022NN}, we focus on the classification according to uncorrelated and correlated noise within a population of neurons. For both types, the noise values will be different over time $t$. Here, we are referring to time only for ESN case, but for FNN it can be associated with different input images.

The separation of noise types occurs depending on how the noise affects the layer with neurons (reservoir for ESN or hidden layer for FNN). If all neurons within one layer receive the same noise value, then we will call such noise correlated and denote it using the superscript `C': $\xi^C(t)$. If all neurons $i$ within one layer receive different noise values, then this is uncorrelated noise $\xi^U(t,i)$. Thus, we consider in total four types of noise: correlated additive $\xi^C_A(t)$ and multiplicative $\xi^C_M(t)$ noise, and uncorrelated additive $\xi^U_A(t,i)$ and multiplicative $\xi^U_M(t,i)$ noise controlling by the noise intensities $D^C_A$, $D^C_M$, $D^U_A$, $D^U_M$, respectively. All this information is illustrated schematically in Fig.~\ref{fig:scheme}(c).

\section{Noise in FNN} \label{sec:noise_FNN}

First we consider the impact of internal noise on FNN when the noise is introduced into the hidden layer of network in accordance with Fig.~\ref{fig:scheme}. In our previous works we have considered how noise impacts on trained FNNs \cite{Semenova2022NN,Semenova2024Chaos}. Here, we are mainly focused on the training process. In this section, we will introduce different noise types during training process and compare its effect on testing performance under noise conditions.

\subsection{Uncorrelated noise in FNN} \label{sec:noise_FNN_uncorr}

\begin{figure}[h]
\includegraphics[width=\linewidth]{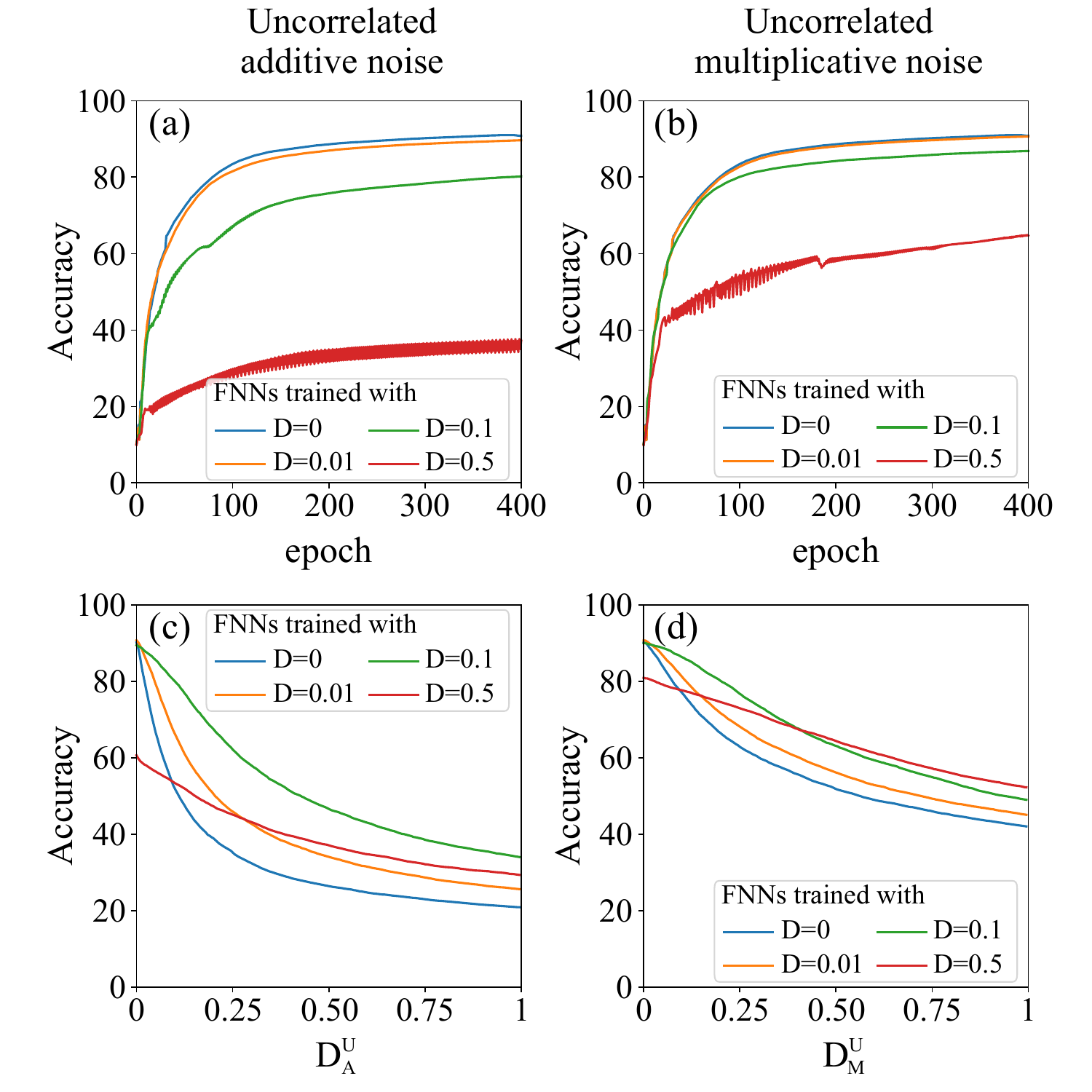}
\caption{\label{fig:FNN_uncorr} Impact of uncorrelated noise on FNN. Left panels show how additive noise of different intensities impacts on the network during training (a) and how the accuracy of FNN trained with four noise intensities $D^U_A=0$, $0.01$, $0.1$, $0.5$ is changed on the testing data when the intensity of uncorrelated additive internal noise is varied (c). The right panels (b,d) were prepared in the same way but for uncorrelated multiplicative noise with $D^U_M=0$, $0.01$, $0.1$, $0.5$.}
\end{figure}

Figure~\ref{fig:FNN_uncorr} shows how uncorrelated noise impacts during training for additive (a) and multiplicative (b) noise. These plots show the dependencies of accuracy of digits recognition depending on the epoch of training. The training was started from the same initial conditions for all four networks. We also ran the same studies with different initial conditions. This resulted in quantitative changes only, not qualitative ones.

The blue curves in Fig.~\ref{fig:FNN_uncorr}(a,b) were prepared for noise free FNN. The rest orange, green and red curves were obtained for FNNs with internal noise during training with intensities $D=0.01$, $D=0.1$, $D=0.5$, respectively. As can be seen from these panels, only noise with large noise intensities can lead to pronounced decrease in accuracy. Otherwise, the result is quite trivial -- the higher the noise intensity, the lower the final accuracy after training.

A more interesting case is when testing a network trained with noise. Figure~\ref{fig:FNN_uncorr}(c,d) shows the performance of all four trained FNNs on testing data when the noise intensity is varied during testing. The color scheme corresponds to the same FNNs as in top panels. The most interesting result is that network trained without internal noise is less resistant to noise then others. Moreover, if network is trained with some large noise intensity, then testing with less noise intensity leads to better performance. This effect is most pronounceable for red curves corresponding to FNNs trained with $D=0.5$. For additive noise, the training accuracy stopped at $\approx 35\%$ (a) and when during testing this intensity was decreased, the accuracy grew up to $\approx 60\%$ (c). The similar effect was obtained for multiplicative uncorrelated noise, when accuracy grew up from $\approx 65\%$ (b) to $\approx 80\%$ (d).

If we ignore the red lines in Fig.~\ref{fig:FNN_uncorr}, we can see a clear tendency that internal noise during the training process helps the network become more noise-resistant. Moreover, if we take a network trained with a relatively high noise intensity of $0.1$ (green lines), its final training accuracy is noticeably lower than for the network trained without noise (see the upper panels in Fig.~\ref{fig:FNN_uncorr}), but if we look at the lower panels, we can see that during testing in the absence of noise, the networks have the same accuracy as noise free trained network, close to $90\%$. If we start to increase the noise intensity during testing, the green curve will be located above the others. Thus, the accuracy of the network trained with noise becomes close to the network trained without noise when testing noise is turned off. At the same time, this network becomes more resistant to noise exposure. The same can be said for the network trained with $D=0.01$ (orange curves).

\subsection{Correlated noise in FNN} \label{sec:noise_FNN_corr}

\begin{figure}[h]
\includegraphics[width=\linewidth]{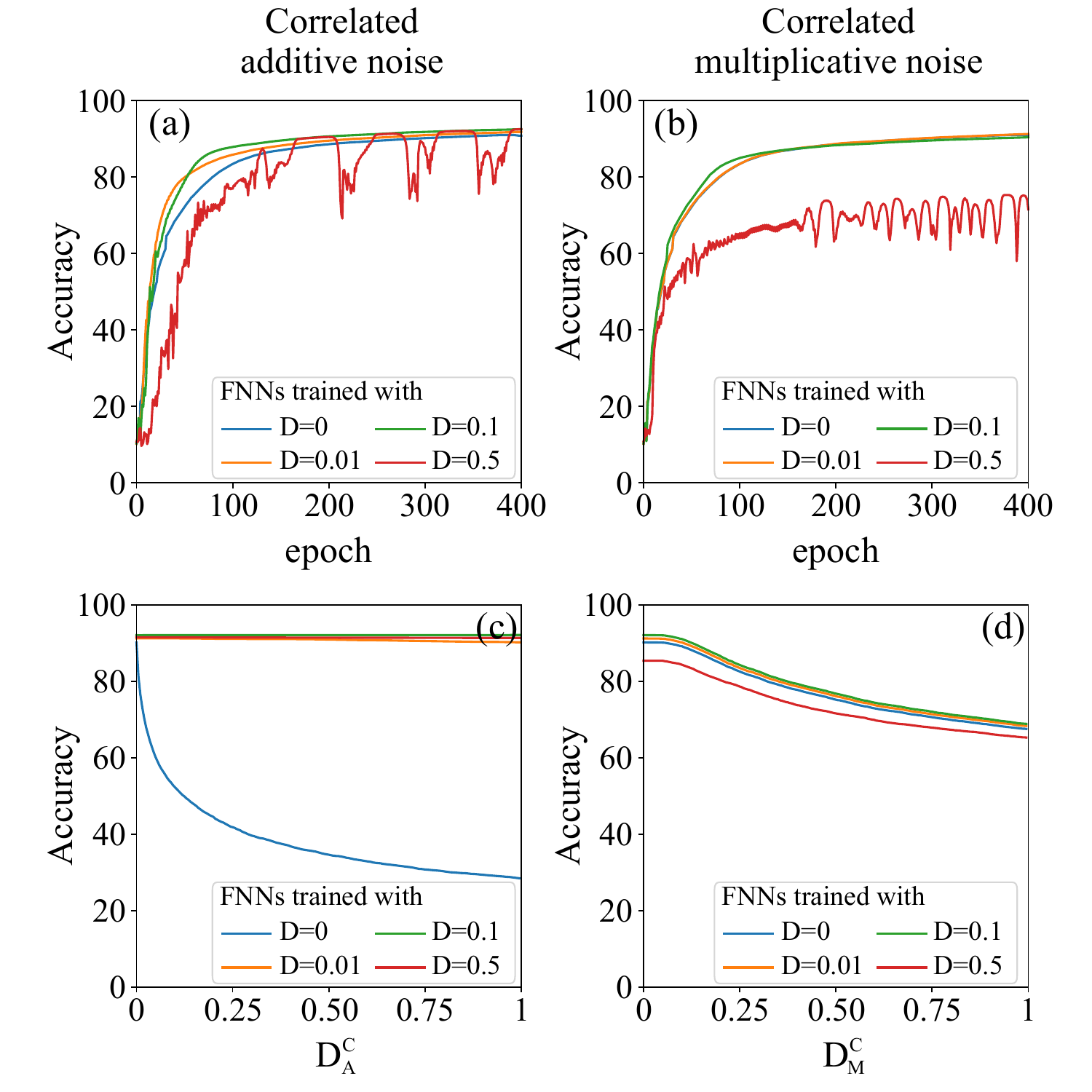}
\caption{\label{fig:FNN_corr} Impact of correlated noise on FNN. Left panels show how additive noise of different intensities impacts on the network during training (a) and how the accuracy of FNN trained with four noise intensities $D^C_A=0$, $0.01$, $0.1$, $0.5$ is changed on the testing data when the intensity of correlated additive internal noise is varied (c). The right panels (b,d) were prepared in the same way but for correlated multiplicative noise with $D^C_M=0$, $0.01$, $0.1$, $0.5$.}
\end{figure}

Figure~\ref{fig:FNN_corr} was prepared in the same way as Fig.~\ref{fig:FNN_uncorr}, but for correlated additive and multiplicative noise. The networks were trained without noise (blue curves) and with noise of intensities $D=0.01$, $0.1$ and $0.5$ (orange, green and red curves, respectively). The left panels in Fig.~\ref{fig:FNN_corr} correspond to additive correlated noise, while the right panels were prepared for multiplicative correlated noise. 

During training, additive correlated noise (see Fig.\ref{fig:FNN_corr}(a)) lead to faster learning and results in slightly more accuracy at the end. The only exception is noise with a relatively very high intensity (red lines). In this case, although additive noise leads to greater accuracy, the training process itself is much slower.

As for the multiplicative correlated noise, the training with large noise slows down training and makes the final accuracy much lower. If the intensities are relatively small, the presence of noise leads to faster learning and almost the same accuracy as without noise.

The bottom panels in Fig.~\ref{fig:FNN_corr} show how the accuracy is changed for all four trained FNNs when noise of different intensities is introduced during testing. These panels were prepared for additive (c) and multiplicative (d) correlated noise. 

Of all the noise types considered, the most obvious difference between networks trained with and without noise can be obtained for additive correlated noise Fig.~\ref{fig:FNN_corr}(c). The graph shows that the network trained without noise (blue curve) has much lower accuracy. As for the networks trained with noise, they become completely resistant to noise, and their accuracy practically does not change even if one introduce noise of greater intensity than that used during training.

As for the multiplicative correlated noise during testing (Fig.~\ref{fig:FNN_corr}(d)), it generally affects the accuracy of the network less than other types of noise. Networks trained without noise and with a small noise intensity behave almost identically, and the curve for FNN trained with large noise intensity $D^C_M=0.5$ is located slightly lower.

\section{Noise in ESN} \label{sec:noise_ESN}
In our previous work \cite{Semenova2025echo}, we have studied how noise impacts on trained ESN depending on noise type, activation function, and etc. Here we are mainly focused on the training process and on comparing the testing performance under noise influence for a network trained without noise and with it.

In the case of noise in recurrent neural network predicted chaotic time series, noise plays a critical role, since even small noise values can lead to unpredictable results. Therefore, for the ESN, lower noise intensities will be considered than for the FNN. It is also important to note that in this section, we will not consider the accuracy of the network, but the RMSE error values (\ref{eq:rmse}). Since the original signal of the Mackey--Glass system belongs to the range of $0.4-1.3$, an error of about $0.5$ can be interpreted as a complete loss of the useful signal accompanied with only a noisy signal from ESN.

In this section, the RMSE will be calculated for a test sequence consisting of 100 points. Since this sequence is chaotic, even a small noise can lead to a wide range of RMSE values depending on initial conditions and noise realization. To maintain the generality of the conclusions, the RMSE will be calculated $K = 100$ times for each noise intensity, and the obtained RMSE values are presented in the form of box plots (see Figs.~\ref{fig:ESN_uncorr},\ref{fig:ESN_corr}).

\subsection{Uncorrelated noise in ESN} \label{sec:noise_ESN_uncorr}

\begin{figure}[h]
\includegraphics[width=\linewidth]{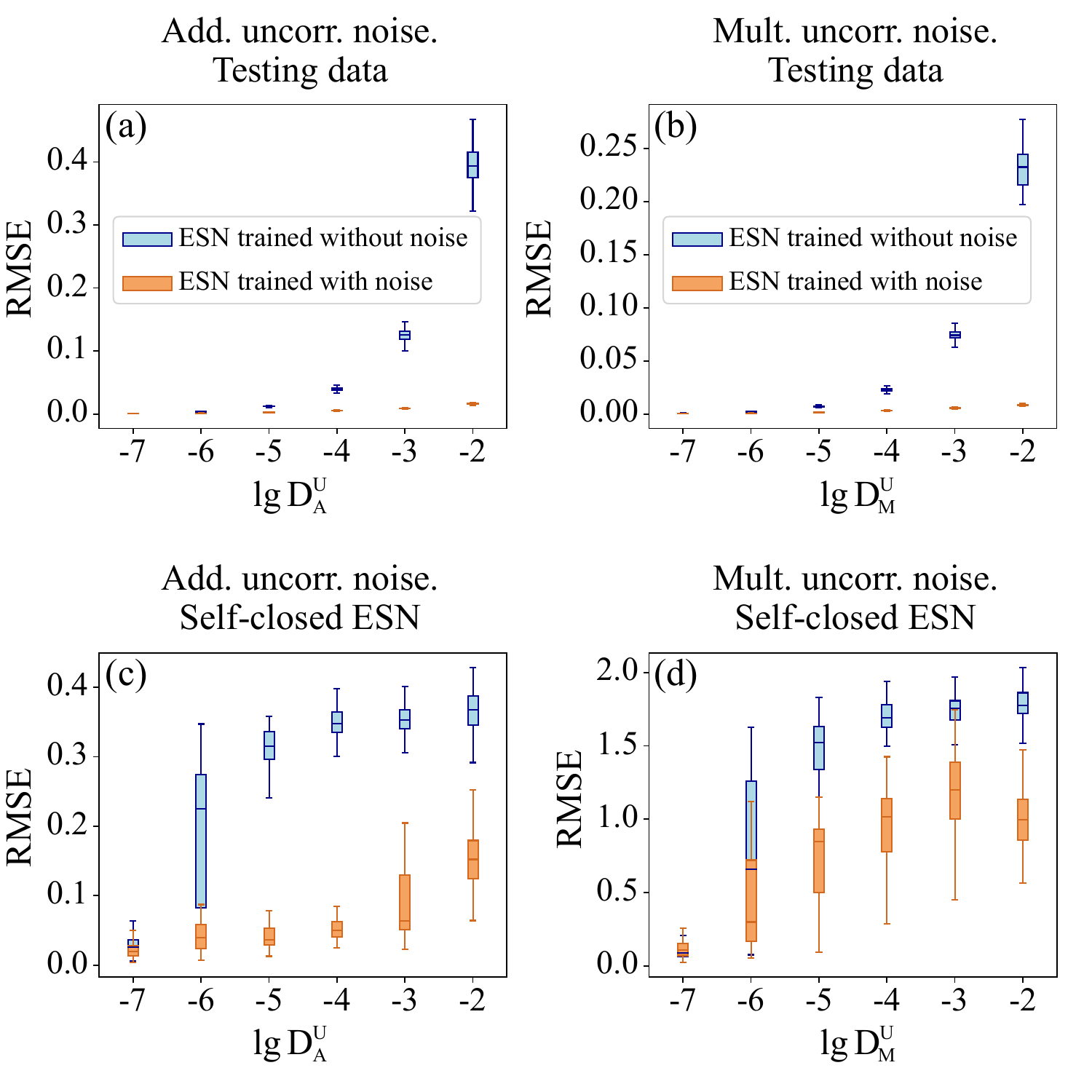}
\caption{\label{fig:ESN_uncorr} Impact of uncorrelated noise on ESN. Left panels show how additive noise of different intensities impacts on the network's RMSE during testing when ESN is not self-closed (a) and when it is self-closed (c). Blue box plots were prepared for ESN trained without noise, orange box plots were obtained for ESNs trained with noise of the same intensities that are used during testing $D^U_A=10^{-7}$, $10^{-6}$, $10^{-5}$, $10^{-4}$, $10^{-3}$, $10^{-2}$. This means that a total of seven ESNs were trained. The right panels (b,d) were prepared in the same way but for uncorrelated multiplicative noise $D^U_M=10^{-7}$, $10^{-6}$, $10^{-5}$, $10^{-4}$, $10^{-3}$, $10^{-2}$.}
\end{figure}

As a first step, let us consider the impact of internal uncorrelated additive and multiplicative noise introduced into reservoir (see Fig.~\ref{fig:ESN_uncorr}). For this purpose we trained one ESN without internal noise and 6 ESNs with additive uncorrelated noise of intensities $10^{-7}$, $10^{-6}$, $10^{-5}$, $10^{-4}$, $10^{-3}$, $10^{-2}$ (left panels of Fig.~\ref{fig:ESN_uncorr}). And the same number of ESNs with the same intensities was trained for multiplicative uncorrelated noise (right panels). It is important to note that all these ESNs were trained from the same initial conditions. We also used the others initial conditions, but this led to only quantitative change. Therefore, in this section the initial conditions are fixed for comparison purposes. These trained networks were applied to testing data when ESNs had the same internal noise intensities as during training.

In the top panels of Fig.~\ref{fig:ESN_uncorr}, blue box plots show how internal noise during testing changes RMSE in not self-closed ESN. If ESNs have been trained with the same noise intensities, then their testing RMSE goes down much lower (orange box plot) than RMSE for the ESN trained without noise (blue box plot). This result holds for both additive (Fig.~\ref{fig:ESN_uncorr}(a)) and multiplicative (Fig.~\ref{fig:ESN_uncorr}(b)) uncorrelated noise.

If ESN becomes self-closed, then its output is sent to the input at next time step. This is expectedly accompanied by the fact that in addition to the noise that comes from the previous values of the reservoir neurons, the noise is also contained in the input signal. The bottom panels in Fig.~\ref{fig:ESN_uncorr} show the impact of additive (c) and multiplicative noise (d) on testing data for this case. The multiplicative noise in self-closed ESN for training both with and without noise lead to similar performance. Already at intensity $D^U_M=10^{-6}$, the error rises higher than $0.5$ indicating a total loss of useful signal. 

The impact of additive uncorrelated noise during the training process has more positive effect on the error (Fig.~\ref{fig:ESN_uncorr}(c)). The complete loss of signal in self-closed ESN trained without noise happens already at $D^U_A=10^{-6}$ during testing. At the same time for networks trained with corresponding noise intensities this happens at much higher intensities starting from $10^{-2}$.

\subsection{Correlated noise in ESN} \label{sec:noise_ESN_corr}

\begin{figure}[h]
\includegraphics[width=\linewidth]{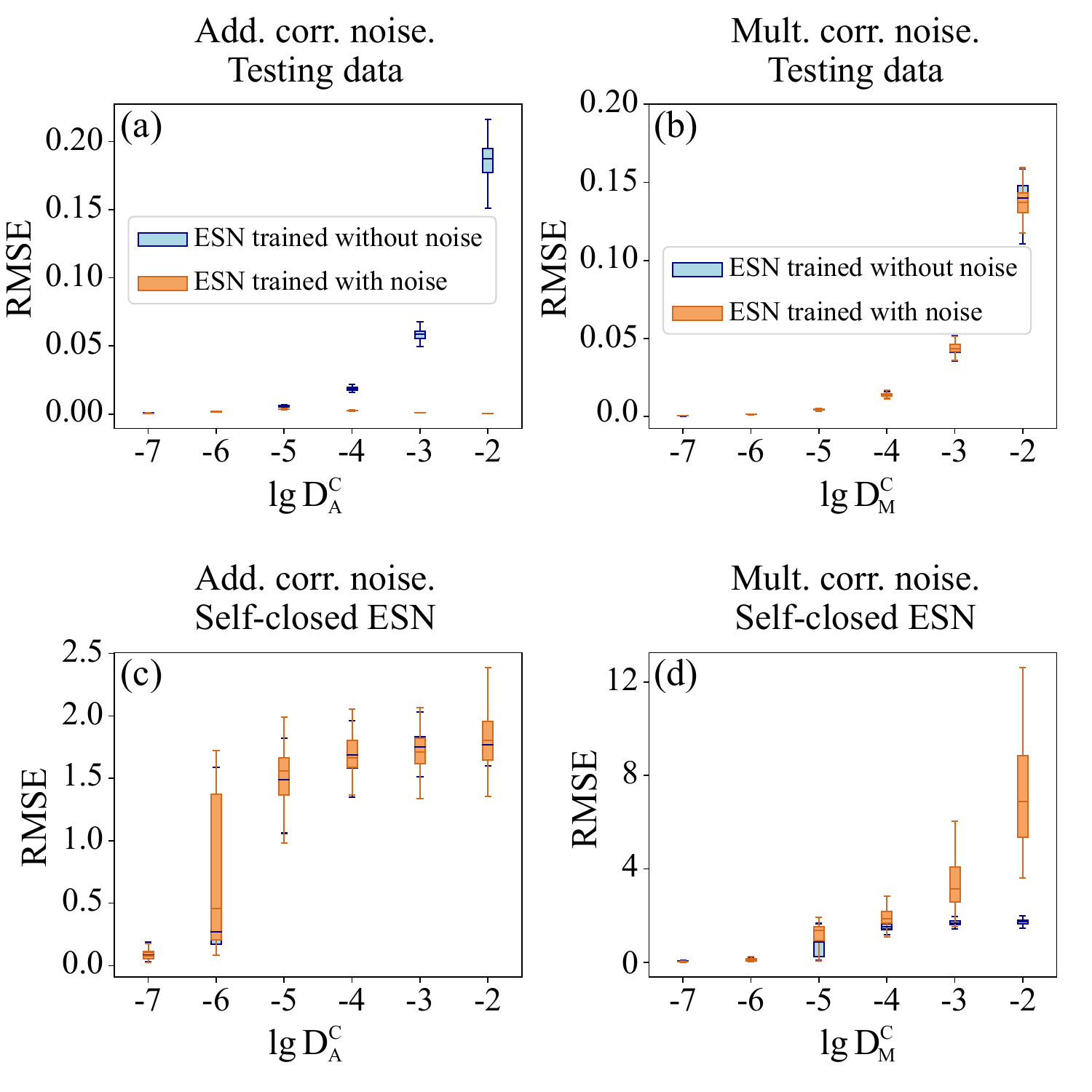}
\caption{\label{fig:ESN_corr} Impact of correlated noise on ESN. Left panels show how additive noise of different intensities impacts on the network's RMSE during testing when ESN is not self-closed (a) and when it is self-closed (c). Blue box plots were prepared for ESN trained without noise, orange box plots were obtained for ESNs trained with noise of the same intensities that are used during testing $D^C_A=10^{-7}$, $10^{-6}$, $10^{-5}$, $10^{-4}$, $10^{-3}$, $10^{-2}$. This means that a total of seven ESNs were trained. The right panels (b,d) were prepared in the same way but for correlated multiplicative noise $D^C_M=10^{-7}$, $10^{-6}$, $10^{-5}$, $10^{-4}$, $10^{-3}$, $10^{-2}$.}
\end{figure}

Correlated noise is more critical for ESNs. Figure~\ref{fig:ESN_corr} shows the results of testing the networks trained without noise and with correlated additive (a,c) and multiplicative (b,d) noise of six intensities. This figure was prepared in the same way as Fig.~\ref{fig:ESN_uncorr}. Top panels show the result of testing of not self-closed ESNs, while the bottom ones were prepared for self-closed ESNs tested on 100 points of Mackey--Glass system's realization. 

As can be seen from Fig.~\ref{fig:ESN_uncorr}(a), training with additive correlated noise helps for testing of not self-closed ESN. The error box plot for internal noise in ESN trained without noise lies much higher than the box plot for networks trained with corresponding noise intensities. Unfortunately, this tendency does not hold for a self-closed ESN (Fig.~\ref{fig:ESN_uncorr}(c)). In this case, networks trained without noise and with noise perform equally poorly. Noise with an intensity of -6 is already enough to completely lose the useful signal.

The most critical noise for the network is multiplicative correlated noise. Its presence does not help in the learning process, and the error when testing a non-self-closed ESN trained without noise and with noise is absolutely the same (Fig.~\ref{fig:ESN_uncorr}(b)). As for the self-closed ESN, here the multiplicative correlated noise during the training process worsens performance, and the error of the network trained with noise is much higher on testing data. At the same time, the error larger than $0.5$ can be interpreted as complete loss of the useful signal. And overcoming this threshold occurs in the same way for $D^C_M=10^{-5}$ both for the network trained with noise and without.

\section{Conclusions}\label{sec:conclu}
In this paper, we have studied how internal noise in neural network during training affects the final quality of the recurrent and deep neural networks. Feedforward (FNN) and echo state (ESN) networks were considered as examples.

In most cases for both deep and echo networks, internal noise during testing helps to improve their resistance to noise, and testing performance for the same noise intensities becomes much higher for networks trained with noise than without.

Deep networks trained with noise perform better for the same internal noise intensity that was used during training, but sometimes even for larger ones. Using uncorrelated noise and correlated multiplicative noise during training reduces the final training accuracy of the network. The higher the noise intensity, the lower the final accuracy. At the same time, if one then reduce the noise intensity during testing compared to what was used during training, the accuracy of the network will increase.

Using additive correlated noise in a deep network during training allows one to almost completely remove sensitivity to this type of noise in the trained network. In an echo network, it allows one to reduce noise on testing data, but the accuracy of a self-closed echo network does not differ in any way when trained without or with noise.

Multiplicative correlated noise during training has almost no effect on the quality of both deep and recurrent networks. Deep network and net self-closed ESN trained without noise and with noise behave very similarly. Multiplicative correlated noise in a self-closed ESN leads to a significant increase in error. However, the loss of useful signal occurs at the same noise intensity for both the network trained without noise and with noise.

In a recurrent network, the impact of uncorrelated additive noise can be significantly reduced if noise is added during training. Even self-closed echo networks perform slightly better. Correlated noise is more critical for recurrent networks. Using additive noise during training significantly reduces noise on the test data, but the difference is erased for a self-closed network.

%\section*{Supplementary Material}
%Our supplementary material contains a pdf-file with additional description of noise reduction methods and how they can be implemented using Python code. It starts with pooling technique and then we consider all three ghost neurons.

\begin{acknowledgments}
This work was supported by the Russian Science Foundation (project No. 23-72-01094) %\hyperref{https://rscf.ru/project/23-72-01094/}
\end{acknowledgments}

\section*{Data Availability Statement}
The data that support the findings of this study are available from the corresponding author upon reasonable request.

\section*{References}
%\bibliography{bibliography}% Produces the bibliography via BibTeX.

\end{document}